\documentclass[runningheads]{llncs}

\usepackage{eccv}

\usepackage{eccvabbrv}

\usepackage{graphicx}
\usepackage{booktabs}

\usepackage[accsupp]{axessibility}  % Improves PDF readability for those with disabilities.

\usepackage{hyperref}

\usepackage{orcidlink}

\usepackage{amsmath,amssymb,amscd,mathrsfs}
\usepackage{amsfonts,bbm}
\usepackage{arydshln}
\usepackage{enumitem}
\usepackage{float}
\usepackage{ifthen}
\usepackage[framemethod=tikz]{mdframed}
\usepackage[bb=dsserif]{mathalpha}
\usepackage{mathtools}
\usepackage{multirow}
\usepackage{pgfplots}
\usepackage{pifont}
\usepackage{placeins}
\usepackage{subcaption}
\usepackage{svg}
\usepackage{tabularx}

\definecolor{shadecolor}{rgb}{0.88,0.93,0.93}

\usepackage[%
  square,        % for square brackets
  comma,         % use commas as separators
  numbers,       % for numerical citations;
  sort&compress, % as sort but in addition multiple numerical citations
]{natbib}

\usepackage[capitalize,noabbrev]{cleveref}

\usepackage{crossreftools}
\usepackage{xargs} 
\usepackage{xstring}
\usepackage{etoolbox}

\usepackage{etoc}
\usepackage[bibliography=common]{apxproof}

\newtheoremrep{thm}{Theorem}[section]
\newtheoremrep{cor}[thm]{Corollary}
\newtheoremrep{lemma}[thm]{Lemma}
\newtheoremrep{prop}[thm]{Proposition}
\newtheoremrep{conj}[thm]{Conjecture}

\crefname{thm}{theorem}{theorems}
\crefname{assumption}{assumption}{assumptions}
\crefname{cor}{corollary}{corollaries}
\crefname{prop}{proposition}{propositions}
\crefname{lemma}{lemma}{lemmas}
\crefname{conj}{conjecture}{conjectures}

\newmdtheoremenv{algo}{Algorithm}
\newlist{assnum}{enumerate}{1} % also creates a counter called 'assnumi'
\setlist[assnum]{label=(\roman*), ref=\theassumption(\roman*)}
\crefalias{assnumi}{assumption} 

\newlist{lemnum}{enumerate}{1} % also creates a counter called 'assnumi'
\setlist[lemnum]{label=(\roman*), ref=\thelemma(\roman*)}
\crefalias{lemnumi}{lemma} 

\newlist{thmnum}{enumerate}{1} % also creates a counter called 'thmnumi'
\setlist[thmnum]{label=(\roman*), ref=\thethm(\roman*)}
\crefalias{thmnumi}{thm} 

\newlist{cornum}{enumerate}{1} % also creates a counter called 'cornumi'
\setlist[cornum]{label=(\roman*), ref=\thecor(\roman*)}
\crefalias{cornumi}{cor} 

\newlist{definitionnum}{enumerate}{1} % also creates a counter called 'definitionnumi'
\setlist[definitionnum]{label=(\roman*), ref=\thedefinition(\roman*)}
\crefalias{definitionnumi}{definition} 

\newlist{conjnum}{enumerate}{1} % also creates a counter called 'conjnum'
\setlist[conjnum]{label=(\roman*), ref=\theconj(\roman*)}
\crefalias{conjnumi}{conj} 

\usepackage{nicefrac}

\usepackage{tcolorbox}
\newtcolorbox{defbox}{colback=black!5!white,colframe=black!75!black}
\newtcolorbox{asmbox}{colback=black!5!white,colframe=black!75!black}
\newtcolorbox{thmbox}{colback=red!5!white,colframe=red!75!black}
\usepackage{braket}

\newcommandx{\QC}[2][1={},2={}]{\ifstrempty{#1}{Q#2}{Q_{#1}\ifstrempty{#2}{}{(#2)}}}
\newcommandx{\PC}[2][1={},2={}]{\ifstrempty{#1}{P#2}{P_{#1}\ifstrempty{#2}{}{(#2)}}}
\newcommandx{\HC}[2][1={},2={}]{\ifstrempty{#1}{H#2}{H_{#1}\ifstrempty{#2}{}{(#2)}}}
\newcommandx{\MC}[2][1={},2={}]{\ifstrempty{#1}{M#2}{M\ifstrempty{#2}{}{(#2)}}}

\newcommandx{\EF}[2][1={k},2={}]{\mathbb E\ifstrempty{#1}{}{_{#1}}#2}

\usepackage[ruled,vlined,linesnumbered]{algorithm2e}
\crefname{algorithm}{Algorithm}{Algorithms}

\title{TMPDiff: Temporal Mixed-Precision for Diffusion Models}

\author{Basile Lewandowski\inst{1}$^*$ \and
        Simon Kurz\inst{2,5}$^*$ \and
        Aditya Shankar\inst{3} \and
        Robert Birke\inst{4} \and
        Jian-Jia Chen\inst{2,5} \and
        Lydia Y. Chen\inst{1,3}}

\authorrunning{B. Lewandowski, S. Kurz et al.}

\institute{
    University of Neuchâtel, Switzerland \\
    \and
    TU Dortmund University, Germany \\
    \and
    TU Delft, Netherlands \\
    \and
    University of Turin, Italy \\
    \and
    LAMARR Institute, Germany
}

\begin{document}

\maketitle
{\renewcommand{\thefootnote}{}\footnotetext{$^*$ Equal contribution.}}
\begin{abstract}
Diffusion models are the go-to method for Text-to-Image generation, but their iterative denoising processes has high inference latency. Quantization reduces compute time by using lower bitwidths, but applies a fixed precision across all denoising timesteps, leaving an entire optimization axis unexplored. We propose TMPDiff, a temporal mixed-precision framework for diffusion models that assigns different numeric precision to different denoising timesteps. We hypothesize that quantization errors accumulate additively across timesteps, which we then validate experimentally. Based on our observations, we develop an adaptive bisectioning-based algorithm, which assigns per-step precisions with linear evaluation complexity, reducing an otherwise exponential search problem. Across four state-of-the-art diffusion models and three datasets, TMPDiff consistently outperforms uniform-precision baselines at matched speedup, achieving 10 to 20\% improvement in perceptual quality. On FLUX.1-dev, TMPDiff achieves 90\% SSIM relative to the full-precision model at a speedup of $2.5\times$ over 16-bit inference. 
\end{abstract}

\section{Introduction}
\label{sec:intro}
Diffusion models are now ubiquitous in generative tasks, with text-to-image (T2I) systems being a leading application. 
As interest in these models has grown, so has their scale, with recent models surpassing ten billion parameters~\cite{flux2024}. 
At inference, this scale compounds with the iterative denoising process of diffusion models. 
Generating an image requires several sequential forward passes through a multi-billion parameter model, making latency the primary limitation~\cite{zeng2025diffusionmodelquantizationreview}. 

A common approach to speed up inference is to use lower bitwidth arithmetic through \textit{quantization}. Early models relied on 32-bit floating point (fp32), but the field quickly shifted to 16-bit (fp16/bf16), with 8- and 4-bit formats now widespread~\cite{zeng2025diffusionmodelquantizationreview}, and even lower precisions under exploration~\cite{zheng2024bidm}. Reduced bitwidth improves speed~\cite{zhao2025viditq}, but it also degrades output quality. Quantization techniques therefore seek the best tradeoff between \textbf{computational efficiency} and \textbf{output fidelity}.

Prior work on diffusion quantization highlighted that quantization errors evolve throughout the denoising process \cite{shang2023ptqdm,he2023ptqd,li2023qdiff}, inspiring \textit{spatial mixed-precision} methods~\cite{zhao2024mixdq, feng2025mpqdm, kim2025mixdit} that assign higher bitwidth to more sensitive model components. While effective, these approaches are fundamentally \textit{static} at runtime: the precision assignment is fixed during calibration and applied uniformly across all diffusion timesteps.

The aforementioned approach is \textit{spatial} quantization, i.e., precision varies only across model components, disregarding varying sensitivities across timesteps.  
We address this gap by introducing the notion of \emph{temporal mixed precision}, which  optimizes precision along the complementary temporal axis across diffusion steps. Concretely, we develop TMPDiff, a novel temporal mixed-precision framework for diffusion models, that assigns different bitwidths to the model at different denoising timesteps. Our approach is guided by three observations. First, per-timestep quantization errors vary significantly in magnitude and impact. Second, these errors accumulate in a predictable, approximately additive manner. Third, this structure enables a principled assignment of precision across timesteps, avoiding costly brute-force search.

Through extensive experiments across four state-of-the-art (SoTA) diffusion models (FLUX.1-dev~\cite{flux2024}, PixArt-$\Sigma$~\cite{chen2024pixart}, Sana~\cite{xie2024sana} and StableDiffusion XL~\cite{sdxl}) and three benchmark datasets (COCO~\cite{lin2014microsoft}, DCI~\cite{urbanek2024picture} and MJHQ~\cite{li2024playground}), we show that TMPDiff consistently outperforms uniform-precision baselines at matched speedups. 
Specifically, TMPDiff yields $10$ to $20\%$ improvement in perceptual quality metrics (SSIM~\cite{ssim}, LPIPS~\cite{zhang2018unreasonable}, PSNR) compared to both full-precision and uniformly quantized pipelines. 
On FLUX.1-dev we achieve excellent fidelity (90\% SSIM) while delivering a speedup of $2.5\times$ over the 16-bit baseline, reducing inference latency by $60\%$ with minimal quality degradation. 
We also show that our temporal precision schedules balance the tradeoffs between the high quality of all full-precision pipelines and the speed of quantized ones.

\medskip
\noindent We summarize \textbf{our contributions} as follows:

\ding{172} \textit{Additive Error Model and Validation}: 
We propose an additive error model for temporal mixed-precision diffusion schedules and empirically validate it across two models and three datasets, demonstrating that per-timestep quantization errors accumulate approximately additively in the final latent (\cref{sec:explorationoftmpeffects}).

\ding{173} \textit{TMPDiff: A Framework for Temporal Precision Scheduling}: 
Building on the additive error model, we propose TMPDiff, a framework with an adaptive bisectioning algorithm that identifies the optimal per-timestep precision assignment with calibration in \textbf{linear} time, reducing an otherwise exponential combinatorical search to a greedy procedure (\cref{sec:algo}).

\ding{174} \textit{Generality of Temporal Sensitivity Scoring}: 
We show that per-timestep sensitivity scoring captures a fundamental property of the denoising process. 
When applied beyond quantization to mixing models of different sizes across timesteps (\cref{subsec:mixofmodels}), TMPDiff outperforms evolutionary search based approaches, demonstrating broader applicability of temporal scheduling.

\section{Background and Problem Definition}
Diffusion models generate data in a \textit{Markovian} fashion by iteratively removing noise \cite{song2021denoising,ho2020denoising}. The model, parameterized by $\theta$, starts from pure Gaussian noise $\mathbf{x}_T\sim \mathcal{N}(0, I_d)$, and applies $T$ denoising steps to recover a clean sample $\mathbf{x}_0 \in \mathbb{R}^d$. At each timestep $t \in \{T,T-1,..,1\}$, the model operates on an intermediate sample $\mathbf{x}_t\in\mathbb{R}^d$. It predicts the noise component $\mu_{\theta}(\mathbf{x}_t,t)\in\mathbb{R}^d$ contained in $\mathbf{x}_t$, and removes it to give the reverse step $\mathbf{x}_t\rightarrow\mathbf{x}_{t-1}$. Under the Denoising Diffusion Implicit Model (DDIM) formulation~\cite{song2021denoising}, the update is deterministic, given by:
\begin{equation}\label{eq:ddim_update}
    \mathbf{x}_{t-1} = \frac{\sqrt{\alpha_{t-1}}}{\sqrt{\alpha_t}} \mathbf{x}_t + B_t \mu_\theta(\mathbf{x}_t, t),
\end{equation}
where $\alpha_t \in (0, 1)$ is a monotonically decreasing noise schedule, and 
\[B_t = \sqrt{1 - \alpha_{t-1}} - \frac{\sqrt{\alpha_{t-1} (1 - \alpha_t)}}{\sqrt{\alpha_t}}.\]

Quantization discretizes the model parameters $\theta$ into $\hat{\theta}$, introducing a local error $\boldsymbol{\delta}_t$ at each denoising step. 
We decompose the quantized noise prediction into the full-precision prediction and a local additive error term $\boldsymbol{\epsilon}_t \in \mathbb{R}^d$ with $\mu_{\hat{\theta}}(\mathbf{x}_t + \boldsymbol{\delta}_t, t) = \mu_{\theta}(\mathbf{x}_t + \boldsymbol{\delta}_t, t) + \boldsymbol{\epsilon}_t$. 
Here, $\boldsymbol{\delta}_t \in \mathbb{R}^d$ denotes the cumulative deviation in the latent caused by local errors of all previous denoising steps, with $\boldsymbol{\delta}_T=\mathbf{0}$. 
Inserting the quantized denoising into the DDIM update rule \eqref{eq:ddim_update} with noisy input $\mathbf{x}_t + \boldsymbol{\delta}_t$ yields:

\begin{equation}\label{eq:ddim_update_with_errors}
    \mathbf{x}_{t-1} + \boldsymbol{\delta}_{t-1} = \frac{\sqrt{\alpha_{t-1}}}{\sqrt{\alpha_t}} (\mathbf{x}_t + \boldsymbol{\delta}_t) 
    + B_t (\mu_{\theta}(\mathbf{x}_t + \boldsymbol{\delta}_t, t) + \boldsymbol{\epsilon}_t)
\end{equation}

To obtain a closed-form solution for the accumulated deviation $\boldsymbol{\delta}_{t-1}$ in the output, Liu \etal~\cite{liu2025error} approximate the full-precision model evaluated at the perturbed input $\mu_{\theta}(\mathbf{x}_t + \boldsymbol{\delta}_t, t)$ by linearizing around the "clean" value $\mathbf{x}_t$ via first-order Taylor expansion:

\begin{equation}\label{eq:jacobian_approximation}
    \mu_{\theta}(\mathbf{x}_t + \boldsymbol{\delta}_t, t) = \mu_{\theta}(\mathbf{x}_t, t) + \mathbf{J}_{\theta}\boldsymbol{\delta}_t,
\end{equation}
\noindent where $\mathbf{J}_{\theta} \in \mathbb{R}^{d \times d}$ is the Jacobian of the noise prediction model. Substituting~\eqref{eq:jacobian_approximation} into~\eqref{eq:ddim_update_with_errors} and subtracting the clean DDIM update from~\eqref{eq:ddim_update} eliminates all error-free terms. What remains is a recursion purely over deviations: 

\begin{equation}
    \boldsymbol{\delta}_{t-1} = \frac{\sqrt{\alpha_{t-1}}}{\sqrt{\alpha_t}} \boldsymbol{\delta}_t 
    + B_t (\mathbf{J}_{\theta}\boldsymbol{\delta}_t + \boldsymbol{\epsilon}_t)
\end{equation}

Grouping $\boldsymbol{\delta}_t$ terms into $\mathbf{A}_t = \frac{\sqrt{\alpha_{t - 1}}}{\sqrt{\alpha_{t}}}I + B_t \mathbf{J}_{\theta}$, yields the compact recursion: 

\begin{equation}\label{eq:errorproprecursion}
    \boldsymbol{\delta}_{t-1} = \mathbf{A}_t \boldsymbol{\delta}_t + B_t \boldsymbol{\epsilon}_t
\end{equation}

Recursively unrolling \eqref{eq:errorproprecursion} from timestep $T$ to $0$, assuming $\boldsymbol{\delta}_T=\mathbf{0}$ (no initial cumulative error), yields a closed-form expression for the final deviation:

\begin{equation}\label{eq:total_error_accumulation}
    \boldsymbol{\delta}_{0} = \sum_{t = 1}^T \left( \prod_{j = 1}^{t - 1} \mathbf{A}_j \right) B_t \boldsymbol{\epsilon}_t.
\end{equation}
This analysis holds across other denoising schedules, as shown by \citet[]{liu2025error}. Minimizing the final accumulated error in~\eqref{eq:total_error_accumulation} is the main objective of quantization in diffusion models. However, \eqref{eq:total_error_accumulation} implicitly assumes that \emph{all} timesteps are executed in quantized precision, \ie, that local quantization errors are injected at every reverse step. But inference can be performed under a \emph{mixed-precision} regime, where only a subset of timesteps are quantized while others remain in full precision, yielding a more general formulation in which quantization errors are injected selectively across time.
\\\\
\textbf{Problem definition.}
Let $Z \in \{0,1\}^T$ denote a \emph{temporal precision schedule}, where $z_t = 1$ indicates full-precision execution and $z_t = 0$ indicates quantized execution at timestep $t$. The schedule $Z$ determines at which timesteps local quantization errors are introduced into the recursive deviation dynamics in~\eqref{eq:errorproprecursion}.

Let $\boldsymbol{\delta}_0(Z)$ denote the final deviation induced by executing the reverse process under schedule $Z$. When $z_t = 0$ for all $t$, i.e., $\boldsymbol{\delta}_0(\mathbf{0})$, it reduces to the fully-quantized case in ~\eqref{eq:total_error_accumulation}. More generally, different schedules induce different accumulated deviations due to the recursive propagation of selectively injected local errors. Let $K = |Z|$ denote the number of full-precision timesteps, which directly controls the inference latency budget. We define the \emph{Temporal Mixed-Precision (TMP)} problem as finding a precision schedule $Z^{\ast}$ that minimizes the final deviation under a budget of $K$ full-precision steps:

\begin{equation}
\label{eq:tmpdefn}
    Z^{\ast} = \arg\min_{Z: |Z| = K} \big\| \boldsymbol{\delta}_0(Z) \big\|
\end{equation}

\section{Additive Error Model and Empirical Analysis}
\label{sec:explorationoftmpeffects}
 
 We first propose an additive ansatz (\cref{subsec:additive_model}) to model final deviations under arbitrary temporal precision schedules. We then empirically validate this additive model in \cref{subsec:singlestep,subsec:maineffect}, demonstrating that it accurately captures error propagation in mixed-precision diffusion and provides a tractable foundation for optimizing the TMP schedule. 

\subsection{Additive Error Ansatz for Mixed Precision Schedules}
\label{subsec:additive_model}
To solve the TMP optimization problem defined in~\eqref{eq:tmpdefn}, 
we need an analytical model of $\boldsymbol{\delta}_0(Z)$ for arbitrary 
temporal precision schedules. The formulation in \eqref{eq:total_error_accumulation} proposed in \cite{liu2025error} 
implicitly assumes that $z_t = 0$ for all $t$, i.e., that all steps are fully quantized. 
How should we modify this if some steps are executed in full precision and others quantized?
We take a natural approach, and propose the following additive \textit{ansatz}; essentially a conjectured extension:

\begin{equation}
\boldsymbol{\delta}_0(Z)
=
\sum_{t=1}^{T}
\left(
\prod_{j=1}^{t-1} \mathbf{A}_j
\right)
B_t \boldsymbol{\epsilon}_t \,(1 - z_t).
\label{eq:additive_ansatz}
\end{equation}

Equation \eqref{eq:additive_ansatz} models the final deviation as a weighted sum of propagated per-timestep 
error contributions, \textit{gated} by the precision schedule. 
As a sanity check, our additive ansatz captures both extremes: setting $z_t = 0$ for all $t$ recovers the fully-quantized case of \eqref{eq:total_error_accumulation}, while setting $z_t = 1$ for all $t$ corresponds to the fully-precise case, where no quantization error is injected.

In the following two subsections, to support the proposed additive model in~\eqref{eq:total_error_accumulation}, we empirically 
validate how it captures 
the behavior of $\boldsymbol{\delta}_0(Z)$ under mixed precision.

\subsection{Validation 1: Single Timestep Error Contributions }
\label{subsec:singlestep}

The additive model predicts that a quantized single timestep $t$ contributes to the final deviation $\boldsymbol{\delta}_0$ by the $t$-th addend in~\eqref{eq:total_error_accumulation}: 

\begin{equation}\label{eq:deltaofsinglet}
    \boldsymbol{\delta}^{(t)}_0 = \left( \prod^{t-1}_{j=1} \mathbf{A}_j \right) B_t \boldsymbol{\epsilon}_t
\end{equation}

To test this simple case first, we change the precision for a single timestep $t$ while keeping all others fixed and measure the resulting change in the final output $\mathbf{x}_0$.
To this end, we define $E(Z)$ as a general measure of the deviation of the final output $\mathbf{x}_0$ under any mixed-precision schedule $Z$ from the full-precision reference by condensing the deviation vector into a scalar via its $L_2$-norm: 

\begin{equation}
    \label{eq:EofZ}
    E(Z) = \|\mathbf{x}^{(Z)}_0 - \mathbf{x}^{(\mathbf{1})}_0 \|_2
\end{equation}
The change of a single timestep can be done in two ways: \emph{upcasting} only a timestep to full-precision, starting from all timesteps being quantized, or \emph{downcasting} by only quantizing a timestep, starting from all timesteps being full-precision.
We use $E(Z)$ to define the \textit{upcasting gain} $\Delta^{\uparrow}_t$ and \textit{downcasting loss} $\Delta^{\downarrow}_t$ as the change in deviation resulting from a single precision change at timestep $t$ accordingly: 

\begin{equation}
    \Delta^\uparrow_t = E(\mathbf{0}) - E(\mathbf{e}_t), \quad 
    \Delta^\downarrow_t = E(\mathbf{1} - \mathbf{e}_t) - E(\mathbf{1}),
\end{equation}
with $\mathbf{0}$ denoting the fully quantized schedule, $\mathbf{1}$ the full-precision schedule, and $\mathbf{e}_t$ the one-hot encoded schedule with full-precision used only for timestep $t$. 
That is, $\Delta^\uparrow_t$ measures the deviation reduction gained by upcasting timestep $t$ to full-precision in an otherwise fully quantized schedule. 
$\Delta^\downarrow_t$ conversely measures the deviation increase by quantizing only timestep $t$ in an otherwise full-precision schedule. 

We perform the following correlation analysis between $\Delta^\uparrow_t$ and $\Delta^\downarrow_t$:
\begin{itemize}
    \item Linear correlation: Pearson $r$~\cite{pearson1895noteonregression} and $R^2$~\cite{draper1998applied} directly test the linearity assumption of our additive model, where a value of $1$ indicates a perfectly linear fit and $R^2$ additionally quantifies the fraction of variance in $E(Z)$ explained by the predictor. 
    \item Rank-order agreement: Spearman $\rho$~\cite{spearman2010theproof} and Kendall $\tau$~\cite{kendall1938anewmeasure} measure rank-order agreement, the property that directly governs schedule selection, while being robust to monotonic nonlinearities and outliers. Kendall $\tau$ further admits a direct probabilistic interpretation via $P = (\tau + 1) \div 2$, where $P$ is the probability that a randomly chosen pair of schedules is ranked correctly by the predictor and $\tau = 0$ corresponds to the 50\% random baseline. 
\end{itemize}

\begin{figure*}[tb]
    \centering
    \begin{subfigure}[tb]{0.495\textwidth}
        \centering
        \includegraphics[width=\linewidth]{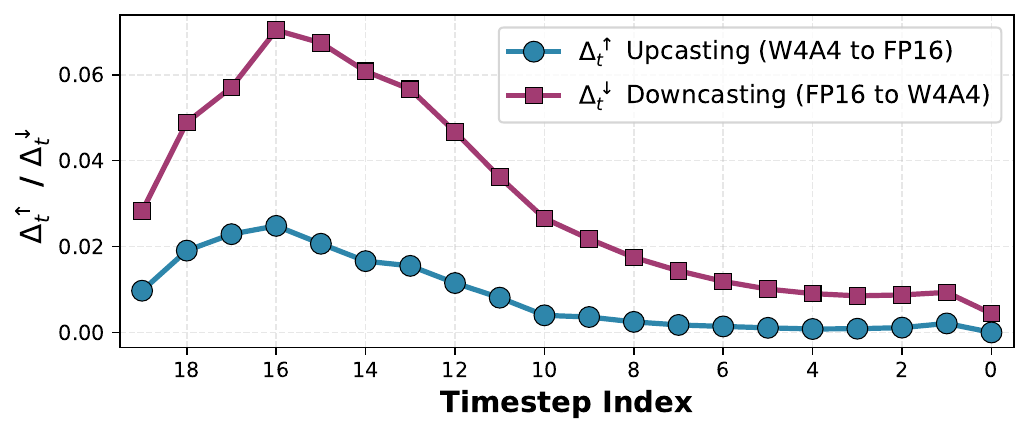}
        \caption{Pixart}
    \end{subfigure}
    \hfill
    \begin{subfigure}[tb]{0.495\textwidth}
        \centering
        \includegraphics[width=\linewidth]{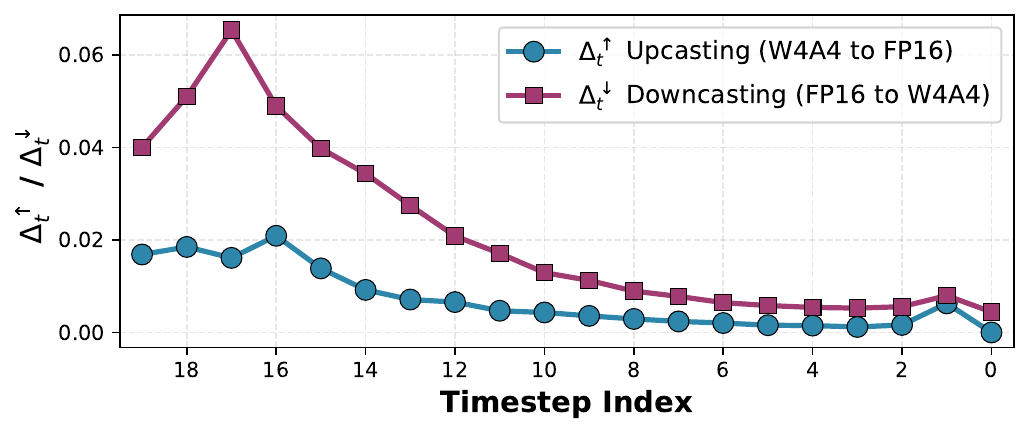}
        \caption{Sana}
    \end{subfigure}
    \caption{Change in final latent error $E(Z)$ from single-timestep up- and downcasting.}
\label{fig:output_errors_w4a4_default}
\end{figure*}

In~\cref{fig:output_errors_w4a4_default}, we evaluate $\Delta^\uparrow_t$ and $\Delta^\downarrow_t$ on 128 samples from the MJHQ dataset~\cite{li2024playground}, using Sana~\cite{xie2024sana} and PixArt-$\Sigma$~\cite{chen2024pixart} with $T=20$ denoising steps and SVDQuant~\cite{li2025svdquant} for quantization of weights and activations to 4-bit (denoted as, W4A4). 
The two curves for each model in~\cref{fig:output_errors_w4a4_default} follow similar trends across timesteps. 
A Pearson $r > 0.94$ and $R^2 > 0.88$ across both models confirms a strong linear relationship between $\Delta^\uparrow_t$ and $\Delta^\downarrow_t$, with over 88\% of the variance of one curve being explained by the other, leaving less than 12\% as residual. Critical for schedule scoring, Spearman $\rho > 0.97$ and Kendall $\tau > 0.88$, imply that over 94\% of random pairwise rankings are correct, confirming that rankings are consistent. 

Together, these results support that per-timestep errors captured by $\Delta^\uparrow_t$ and $\Delta^\downarrow_t$ accumulate additively up to a ranking-neutral residual, indicating that the additive error model holds even when evaluated at the extreme reference points of a fully quantized and a complete full-precision schedule. 

\subsection{Validation 2: Multistep Error Contributions}
\label{subsec:maineffect}

We now further stress-test whether the approximate additivity observed for single precision changes extends to schedules of multiple precision toggles $K = |Z|$ with $1 < K < T$. 
Since errors from individual timesteps accumulate additively, single-toggle measurements alone should suffice to rank multi-toggle schedules, without requiring direct evaluation of $E(Z)$ for each candidate. 
Specifically, the downcasting error reduction of a schedule $Z$ relative to a fully quantized reference schedule is approximately $\sum_{t=1}^T \Delta^{\uparrow}_t z_t$, where $z_t$ gates the upcasting gains of the selected full-precision timesteps. 
For predicting the error gain from quantized timesteps based on $\Delta^\downarrow_t$, we use the fact that $\sum_{t=1}^T \Delta^{\downarrow}_t (1 - z_t) = \sum_{t=1}^T \Delta^{\downarrow}_t - \sum_{t=1}^T \Delta^{\downarrow}_t z_t$ with $\sum_{t=1}^T \Delta^{\downarrow}_t$ being a constant to express the error gain in terms of $z_t$ instead of $1-z_t$. 
We therefore define the following two scoring functions with sign correction for the lower the better ranking:

\begin{equation}
    S^{\uparrow}(Z) = -\sum_{t=1}^T \Delta^\uparrow_t z_t, \quad S^{\downarrow}(Z) = -\sum_{t=1}^T \Delta^\downarrow_t z_t
\end{equation}

Similar to Section~\ref{subsec:singlestep}, we analyze the measured final latent errors $E(Z)$, reporting Spearman $\rho$, Pearson $r$, $R^2$, and Kendall $\tau$ for each $K$ between the measured error and either $S^{\uparrow}(Z)$ or $S^{\downarrow}(Z)$. The tests are conducted as follows: We directly measure $E(Z)$ of 20 random schedules for each $K \in \{2, 6, 10, 14, 18\}$ using the same 128 MJHQ samples used to compute $\Delta^\uparrow_t$ and $\Delta^\downarrow_t$ and compare them against the schedule scores of $S^{\uparrow}(Z)$ and $S^{\downarrow}(Z)$.

\begin{figure*}[tb]
    \centering
    \begin{subfigure}[b]{0.495\textwidth}
        \centering
        \includegraphics[width=\linewidth]{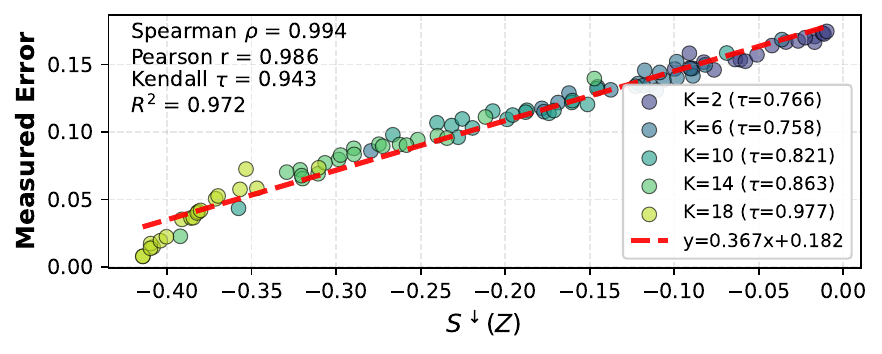}
        \caption{Downcasting Predictor Performance}
    \end{subfigure}
    \hfill
    \begin{subfigure}[b]{0.495\textwidth}
        \centering
        \includegraphics[width=\linewidth]{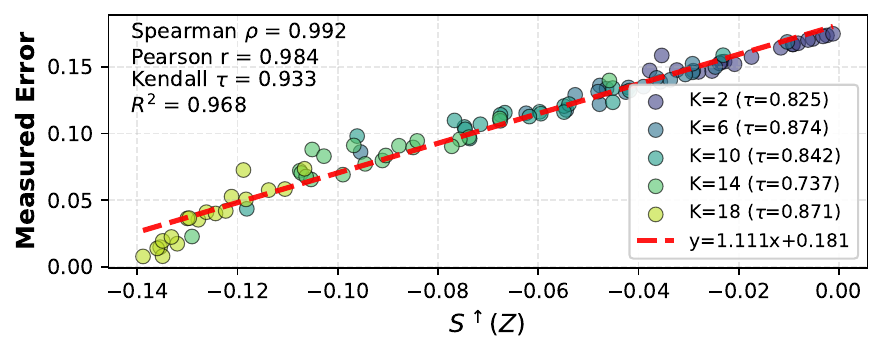}
        \caption{Upcasting Predictor Performance}
    \end{subfigure}
    \caption{Linear fit between the measured final latent error $E(Z)$ and the schedule scoring functions $S^{\uparrow}(Z)$ and $S^{\downarrow}(Z)$ for upcasting $K$ timesteps for Sana.}
    \label{fig:final_error_prediction}
\end{figure*}

\Cref{fig:final_error_prediction} illustrates the results of the above test. 
It shows a strong linear fit between $E(Z)$ and $S^{\uparrow}(Z)$ or $S^{\downarrow}(Z)$ (Pearson $r > 0.98$, $R^2 > 0.96$) and high rank-order agreement (Kendall $\tau > 0.93$, Spearman $\rho > 0.99$), confirming that quantization errors accumulate approximately additively across timesteps. 
Single-toggle measurements reliably predict the ranking of multi-toggle schedules, validating the additive error model for practical schedule search. 

The strong correlations persist on held-out COCO~\cite{lin2014microsoft} and DCI~\cite{urbanek2024picture} dataset samples (Spearman $\rho = 0.99$, Kendall $\tau \geq 0.825$ across all $K$, full results in~\cref{app:outofdistribution}), confirming generalization beyond the calibration distribution. 
Schedules optimized for low mean error also yield lower output variance across diverse prompts, reducing outlier failure risk. 
Results also stay consistent across different amounts of calibration samples, as shown in~\cref{app:calibrationablation}. 
The final output error $E(Z)$ strongly correlates with perceptual image-space metrics as shown in~\cref{app:latenterrorproxy}. Hence, we can use it as a calibration proxy without the large sample sizes that image-space metrics require for stable estimates~\cite{chong2019effectively}. 

\section{TMPDiff: A Mixed-Precision Scheduling Framework}\label{sec:algo}

TMPDiff is a framework that learns an optimized temporal precision schedule offline using calibration data to dynamically switch precision across diffusion timesteps at inference.
\Cref{subsec:speedups} characterizes the quality-latency design space to identify where these schedule decisions have the greatest impact. 
\Cref{subsec:searchspace} provides a theory-grounded sampling strategy that makes optimal schedule search via single-timestep error gains (or losses) tractable for large models. 
\Cref{subsec:tmpdiffpipeline} presents the complete TMPDiff algorithm, including hardware considerations for practical deployment.

\subsection{Targeting High Speedups}
\label{subsec:speedups}

We characterize the latency-design space to make an informed decision on $K$, the number of full-precision denoising steps. 
Targeting an end-to-end speedup $r$ requires at most $K$ full-precision timesteps, describing a hyperbolic relationship governed by Amdahl's law~\cite{gustafson1988reevaluating}: 

\begin{equation}\label{eq:amdahls_law}
    r = \frac{T}{K + \frac{T - K}{\lambda}} \quad \Longleftrightarrow K = \frac{T (\lambda - r)}{r (\lambda - 1)},
\end{equation}
where $T$ is the number of denoising steps and $\lambda$ is a fixed per-timestep speedup from quantization. 
The right-hand side specifies how many full-precision steps $K$ a given total speedup target $r$ permits. 
Crucially, this relationship is hyperbolic in $K$. 
That is, the first few full-precision steps introduced into an otherwise quantized pipeline come at the steepest speedup penalty, while each additional full-precision step costs progressively less. 
Consequently, the regime of small $K$ is where precision allocation decisions matter most, and each full precision timestep must be placed where it recovers the most image quality. 

Since small $K$ corresponds to schedules closer to the fully-quantized baseline, $S^{\uparrow}(Z)$ is the more natural scoring function. 
This is confirmed empirically in~\cref{fig:final_error_prediction}, where $S^{\uparrow}(Z)$ achieves higher Kendall $\tau$ than $S^{\downarrow}(Z)$ for small $K$. 
\Cref{fig:pareto_optimality} confirms that TMPDiff fills the quality-latency between the fully-quantized and full-precision denoising pipelines across the entire range of $K$.

\subsection{Temporal Search Space Exploration}
\label{subsec:searchspace}

Computing the per-timestep upcasting gain $\Delta^{\uparrow}_t$ for schedule scoring requires a full forward pass per timestep, resulting in $T$ calibration evaluations in total.
For small models this is acceptable, for large multi-billion parameter diffusion models and long denoising pipelines it becomes a practical bottleneck. 
Exhaustive evaluation is, however, not necessary if the error gain/loss curve can be reconstructed accurately from a small number of samples. 

\Cref{fig:output_errors_w4a4_default} shows that this is indeed the case as the per-timestep error gain/loss curve exhibits two properties that make sparse sampling viable. 
First, it is smooth with a few spikes, allowing reliable interpolation between points. 
Second, timesteps of high gains/losses concentrate in early denoising steps with a global trend towards lower gains or losses at the end. 
Both properties are visible for all models shown in~\cref{fig:output_errors_w4a4_default}, consistent with the structural role of early timesteps in establishing the global image layout during denoising~\cite{hertz2023prompttoprompt}. 

\begin{figure}[tb]
    \centering
    \includegraphics[width=.65\linewidth]{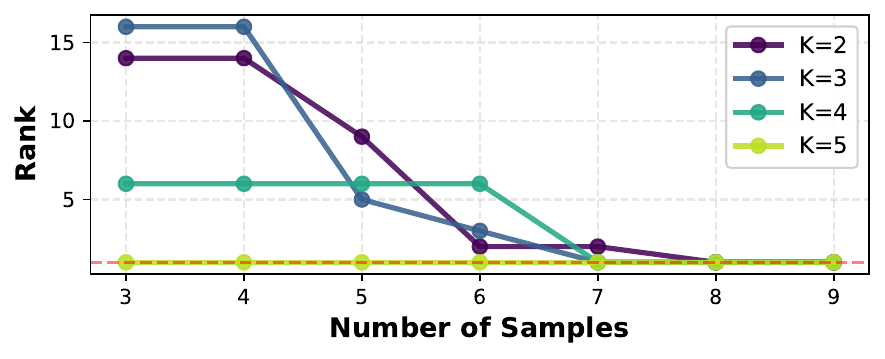}
    \caption{Rank deviation per $K$ over the number of sampled timesteps for calibration on the COCO dataset for the Sana model.}
    \label{fig:convergence}
\end{figure}

These two properties motivate an adaptive bisection strategy that takes targeted measures where the gain/loss curve is highest and interpolates everywhere else. 
We start from the first, middle, and last timestep as three anchor points, and group all unsampled timesteps into segments $[t_L, t_R]$. 
Each segment is scored by the average error gain (or loss) of its already sampled endpoints via $\frac{1}{2} (\Delta^{\uparrow}_{t_L} + \Delta^{\uparrow}_{t_R})$, reflecting the expected error gain of the region it spans. 
We then sample the midpoint of the highest-scoring segment, split it into two new segments and repeat. This procedure is a form of \textit{adaptive bisectioning}. However, unlike binary search, which halves a fixed interval, it refines regions flanked by high upcasting gain, concentrating evaluations on the critical areas of the curve for schedule selection. 

The plausibility of sparse sampling rests on Lipschitz continuity of $\Delta^\uparrow_t$. 
Since the ground-truth upcasting gains $\Delta^{\uparrow}_t$ for timesteps $t\in \{1, ..., T\}$ are finite in magnitude, we can upper-bound the interpolation error between two already sampled timesteps $t_L$ and $t_R$ by a positive Lipschitz constant $L$:
\begin{equation}
    \max_{t \in [t_L, t_R]}| \Delta^{\uparrow}_t - \Delta^{\uparrow}_{t, interp.}| \leq \frac{L}{2} \cdot |t_L - t_R|.
\end{equation}
Refining any segment reduces the global interpolation error
\begin{equation}
\varepsilon = \max_{[t_L, t_R]} \left( \frac{L}{2} \cdot |t_L - t_R| \right). 
\end{equation}
This motivates concentrating evaluations on regions where the gain is highest, rather than sampling uniformly. 
The interpolated top-$K$ timesteps for any number $K$ of full-precision timesteps are provably correct if $2\varepsilon < \Delta_K$ with $\Delta_K \geq 0$ as the difference between the $K$-th and ($K$+1)-th largest $\Delta^{\uparrow}$ of the measurements. 
Intuitively, the error for every interpolated timestep is upper bounded via $\varepsilon$, so in the worst case the $K$-th $\Delta^{\uparrow}$ is underestimated by $-\varepsilon$ while the ($K$+1)-th $\Delta^{\uparrow}$ is overestimated by $+\varepsilon$. 
Then, $2\varepsilon > \Delta_K$ and we confuse $K$ with $K$+1. 
Conversely, once the bisection drives $\varepsilon$ below $\Delta_K/2$, no furhter sampling can change the top-$K$ set of timesteps. 
Importantly, $L$ and $\Delta_K$ are not estimated at runtime, so the Lipschitz bound provides a qualitative justification rather than a computable termination condition. 
In practice, the algorithm runs for a fixed sampling budget $B$ (Algorithm~\ref{alg:tmpdiff}). 

To evaluate convergence, we measure how quickly our sampling recovers the optimal schedule compared to a fully evaluated reference. 
At each iteration, we reconstruct missing error gains/losses via linear interpolation, derive the greedy optimal schedule $\hat{U}^{(K)}$ for target size $K$, and rank it against the top schedules of the fully measured reference. 
\Cref{fig:convergence} shows that starting from three anchor points, the rank deviation drops to zero within fewer than ten iterations across $K \in \{2, 3, 4, 5\}$ on the Sana model for calibration on COCO samples.

\subsection{Adaptive Bisection Algorithm for Precision Scheduling}
\label{subsec:tmpdiffpipeline}

\begin{algorithm}[tb]
    \caption{TMPDiff Algorithm}\label{alg:tmpdiff}
    \KwIn{Full-precision model $F$, quantized model $Q$, 
          calibration data $\mathcal{D}$, denoising steps $T$, 
          target speedup $r$, per-timestep speedup $\lambda$, sampling budget $B$}
    \KwOut{Precision schedule $Z^*$, accelerated image $\mathbf{x}_0$}
    
    $K \leftarrow \lfloor \frac{T(\lambda - r)}{r(\lambda - 1)} \rfloor$ \tcp*{Latency budget \eqref{eq:amdahls_law}}
    
    $\mathcal{A} \leftarrow \{1, \lfloor T/2 \rfloor, T\}$, \quad compute $\Delta^{\uparrow}_t$ $\forall t \in \mathcal{A}$ \tcp*{Anchor init.}
    \While{$|\mathcal{A}| \leq B$ \tcp*{Sparse sampling, \cref{subsec:searchspace}}}{
        $s^* \leftarrow \operatorname*{argmax}_{[t_L, t_R]} \tfrac{1}{2}(\Delta^{\uparrow}_{t_L} + \Delta^{\uparrow}_{t_R})$ \tcp*{Best segment}
        $t^* \leftarrow \lfloor(t_L + t_R)/2\rfloor$, \quad compute $\Delta^{\uparrow}_{t^*}$\;
        $\mathcal{A} \leftarrow \mathcal{A} \cup \{t^*\}$\;
    }
    $\Delta^{\uparrow}_t \leftarrow \text{interpolate}(\mathcal{A})$ \quad $\forall t \notin \mathcal{A}$ \tcp*{Fill gains}
    $Z^* \leftarrow \text{top-}K \text{ indices of } \Delta^{\uparrow}_t$ \tcp*{Greedy schedule}
    
    \For(\tcp*[f]{Inference}){$t = T, \ldots, 1$}{
        $\mathbf{x}_{t-1} \leftarrow F(\mathbf{x}_t, t)$ \textbf{if} $t \in Z^*$ \textbf{else} $Q(\mathbf{x}_t, t)$
    }
    \Return $\mathbf{x}_0$\;
\end{algorithm}

Algorithm~\ref{alg:tmpdiff} outlines the TMPDiff framework, integrating the latency-aware derivation of $K$ of~\cref{subsec:speedups} and the search space exploration of~\cref{subsec:searchspace} into an end-to-end calibration and inference pipeline. 
First, given a latency target, it computes how many timesteps $K$ can run in full precision (line 1).
Exploration is initialized with the first, last and middle diffusion steps as anchors (line 2) before undergoing sparse sampling. We then iteratively sample midpoints of high-gain segments until the calibration budget is exhausted (lines 4-7). 
Then, the gain curve is interpolated across all timesteps, and the top-$K$ greedily-chosen timesteps with the largest gains form the precision schedule (lines 8-9). 

At inference, TMPDiff switches between the full-precision and the quantized model between denoising steps according to the optimal schedule $Z^*$. This can be done in practice through different software systems depending on the hardware resources available, as discussed in \cref{app:memtrans}. 

Crucially, the naive $2^T$ combinatorial search (lines 4–7) collapses to a linear-time procedure: measure each timestep’s single-toggle upcasting gain or downcasting loss, then rank them. 
This reduction follows directly from the additive error model established in~\cref{sec:explorationoftmpeffects}.

\section{Experimental Results}
We first describe our experimental setup in~\cref{subsec:experimental_setup} before evaluating \mbox{TMPDiff} across four diffusion models and three datasets. 
\cref{subsec:pareto_optimality} characterizes the qualtiy-latency design space by empirically evaluating the Pareto frontier over $K$, connecting to the speedup analysis of~\cref{subsec:speedups}. 
We then evaluate image quality against uniform precision baselines at matched speedups in~\cref{subsec:image_quality}, before demonstrating the broader applicability of our temporal scheduling framework beyond quantization in~\cref{subsec:mixofmodels} by mixing models of different sizes.

\subsection{Experimental Setup}
\label{subsec:experimental_setup}

\hspace{\parindent}
\textbf{Metrics}.
Since quantization aims to accelerate inference while preserving generation quality, we use the full-precision model output as reference and report Peak Signal-to-Noise Ratio (PSNR), Structural Similarity Index Measure (SSIM)~\cite{ssim}, and Learned Perceptual Image Patch Similarity (LPIPS)~\cite{zhang2018unreasonable} to measure fidelity alongside achieved speedup. 
Using full-precision model outputs as reference provides a direct and sensitive measure of quantization-induced performance degradation, grounded in well-characterized model behavior. 
Reference-free metrics are less discriminative in this controlled setting but are reported in~\cref{tab:gt_results} of~\cref{app:tables} as corroborative evidence. 
Speedups are measured on our testbed described in~\cref{app:testbed}.

\textbf{Models}. We consider four models of various architectures and sizes in our experiments: FLUX.1-dev~\cite{flux2024}, PixArt-$\Sigma$~\cite{chen2024pixart}, Sana~\cite{xie2024sana} as transformer-based diffusion models and StableDiffusion XL~\cite{sdxl} for its convolutional backbone, with 12B, 0.6B, 1.6B and 3.5B parameters respectively. We use 16-bit floating point formats as default full-precision, either as fp16 for PixArt and SDXL or bf16 for FLUX and Sana.
All images are generated using 20 inference steps and unconditional guidance.

\textbf{Datasets}. We evaluate the models on three datasets: Microsoft Common Objects in COntext (COCO, \cite{lin2014microsoft}), Densely Captioned Images (DCI, \cite{urbanek2024picture}), and MidJourney HQ (MJHQ, \cite{li2024playground}). We use the summarised version of DCI where captions fit within 77 tokens.

\textbf{Quantization.} For quantized inference in TMPDiff, we mainly use SVD-Quant~\cite{li2025svdquant} to produce int4-quantized models. Unless specified otherwise, the quantization setup is 4-bits for weight and activation and a rank-32 correction adapter (W4A4R32\footnote{Throughout this article, we will denote by W$x$A$y$ a quantization method where model weights and activations are quantized to $x$ and $y$ bits, respectively. When SVDQuant is used, we also refer to the adapter rank with R$z$.}). 
As a second quantization technique, we also consider the 8-bit quantized baseline using the TensorRT SDK\footnote{\url{https://developer.nvidia.com/tensorrt}}.

\subsection{Pareto Optimality}
\label{subsec:pareto_optimality}

The varying timestep error gains in~\cref{subsec:singlestep} together with the hyperbolic relationship between the end-to-end speedup $r$ and the number of full-precision denoising steps $K$ of~\cref{subsec:speedups} implies that a small number of well-chosen full-precision steps can recover substantial quality while preserving most of the speedup. 
We verify this empirically by evaluating TMPDiff on FLUX.1-dev for $K=\{4, 8, 12, 16\}$ resulting in speedups of 1.4$\times$ to $2.6\times$ under a total of $T=20$ denoising steps. 

\begin{figure}[tb]
    \centering
    \includegraphics[width=.65\linewidth]{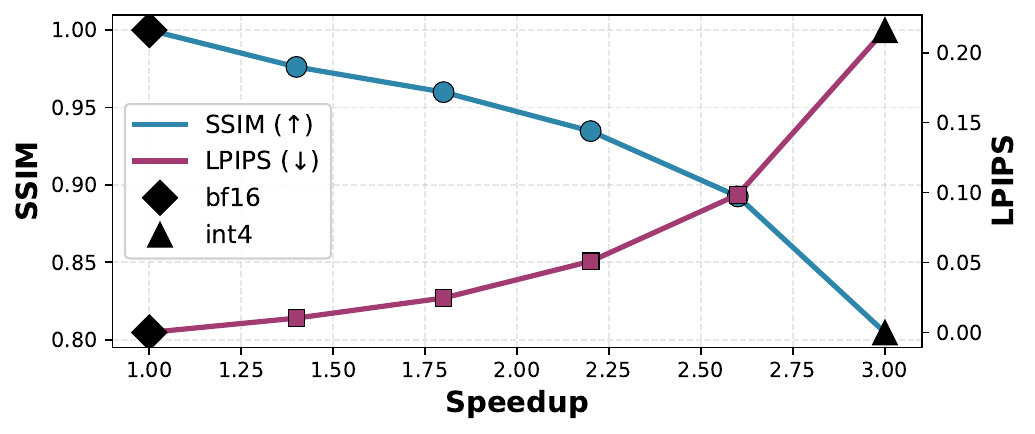}
    \caption{Performance metrics over different speedup factors on FLUX W4A4R32.}
    \label{fig:pareto_optimality}
\end{figure}

\Cref{fig:pareto_optimality} confirms that TMPDiff successfully fills the quality-latency tradeoff gap between the full-precision and uniformly quantized model. 
Every point on the TMPDiff Pareto curve is both faster than the unquantized model and of higher quality than the uniformly quantized one. 
At the high speedup end, a schedule with 80\% quantized denoising steps achieves a significant improvement in image quality (+10 percentage points (pp.) of SSIM \& -12pp. LPIPS) with a limited impact on latency (+27\% \wrt quantized speed). 
At the low speedup end, a schedule with only 20\% quantized steps shows minimal degradation (-3pp. SSIM \& +1pp. LPIPS) with a notable latency reduction (-17\% \wrt initial speed). 
Therefore, TMPDiff allows making deliberate decisions on the tradeoff between quality and speed.

\subsection{Image Quality}
\label{subsec:image_quality}

\begin{table}[tb]
    \centering
    \caption{Comparison across different models and datasets. The reference point for similarity evaluation are the images generated by the unquantized (16 bits) model. To compare models with similar speedups, we scale the number of timesteps depending on the data type.}
    \resizebox{\columnwidth}{!}{
    \begin{tabular}{@{}llcccccccccc@{}}
\toprule
                        \multirow{2}{*}{Model}  & \multirow{2}{*}{Method}           & \multicolumn{3}{c}{MJHQ} & \multicolumn{3}{c}{COCO} & \multicolumn{3}{c}{sDCI} & \multirow{2}{*}{Speedup} \\  \cmidrule(l){3-5} \cmidrule(l){6-8}\cmidrule(l){9-11}
                        &          & SSIM\tiny{$\uparrow$}& PSNR\tiny{$\uparrow$} & LPIPS\tiny{$\downarrow$} & SSIM\tiny{$\uparrow$} & PSNR\tiny{$\uparrow$} & LPIPS\tiny{$\downarrow$} & SSIM\tiny{$\uparrow$} & PSNR\tiny{$\uparrow$} & LPIPS\tiny{$\downarrow$} & \\
\midrule
\multirow{3}{*}{Pixart} & FP16 \small(8 steps)   & 0.58 & 15.3 & 0.41 & 0.61 & 15.4 & 0.40 & 0.53 & 15.1 & 0.41 & 2.5 \\
                        & W4A4 \small(25 steps)  & 0.63 & 16.8 & 0.36 & 0.61 & 15.6 & 0.40 & 0.54 & 15.4 & 0.40 & -- \\
                        & Ours \small(15+5 steps) & \textbf{0.69}                          & \textbf{19.4}                          & \textbf{0.27}                             & \textbf{0.69}                          & \textbf{18.4}                          & \textbf{0.30}                             & \textbf{0.61}                          & \textbf{17.8}                          & \textbf{0.31}                             & --              \\ 
\midrule
\multirow{3}{*}{Sana}   & BF16 \small(8 steps) & 0.54       & 14.6      & 0.38     & 0.53       & 13.5      & 0.41     & 0.45       & 12.0      & 0.44     & 2.5      \\
                        & W4A4 \small(25 steps) & 0.63       & 17.0      & 0.26     & 0.60       & 15.6      & 0.31     & 0.56       & 15.3      & 0.30     & 2.6       \\
                        & Ours \small(15+5 steps) & \textbf{0.70}                          & \textbf{19.2}                          & \textbf{0.19}                             & \textbf{0.70}                          & \textbf{18.8}                          & \textbf{0.20}                             & \textbf{0.64}                          & \textbf{17.6}                          & \textbf{0.20}                             & 2.68                     \\ 
\midrule
\multirow{2}{*}{SDXL}   & TensorRT int8 & 0.42       & 12.2      & 0.67     & 0.46       & 12.4      & 0.69     & 0.38       & 12.5      & 0.69     & 1.24       \\
                        & Ours                                   & \textbf{0.85}                          & \textbf{28.0}                          & \textbf{0.10}                             & \textbf{0.87}                          & \textbf{28.5}                          & \textbf{0.11}                             & \textbf{0.83}                          & \textbf{27.6}                          & \textbf{0.11}                             & 1.33                     \\ 
\midrule
\multirow{2}{*}{FLUX}   & TensorRT fp8 & 0.46       & 11.0      & 0.71     & 0.47       & 10.2      & 0.75     & 0.37       & 9.0      & 0.81     & 1.57       \\
                        & Ours                                   & \textbf{0.91}                          & \textbf{28.1}                          & \textbf{0.08}                             & \textbf{0.91}                          & \textbf{27.4}                          & \textbf{0.07}                             & \textbf{0.87}                          & \textbf{25.3}                          & \textbf{0.10}                             & 2.5                      \\ 
\bottomrule
\end{tabular}
}
\label{tab:results_everything}
\end{table}

We evaluate TMPDiff against uniform precision baselines using a schedule with $75\%$ quantized timesteps (15 out of 20 steps, 15+5). 
To ensure fair comparison, we match speedups across methods by adjusting the number of denoising steps, ensuring that any quality advantage of TMPDiff cannot be attributed to a more favourable latency budget.
Specifically, for PixArt and Sana, we compare against a full-precision model with 8 timesteps (2.5$\times$ speedup) and a uniformly quantized W4A4 SVDQuant model with 25 steps (2.6$\times$ speedup). 
Both reference pipelines achieve similar speedups to our TMPDiff pipeline. 
For FLUX and SDXL we compare against the TensorRT 8-bit quantization baseline.
As SVDQuant does not yet support efficient hardware deployment of PixArt, no speedup is reported. 
Sana provides a representative speedup estimate for PixArt given their architecture similarity. 

\Cref{tab:results_everything} shows that we consistently outperform the uniform precision baselines across all models and datasets. 
TMPDiff achieves strongest results on FLUX and SDXL, with exceptional fidelity throughout all datasets and metrics at $2.5\times$ speedup for FLUX and $1.33\times$ speedup for SDXL. 
The latter result is particularly notable. 
Despite running in plenty 8-bit precision, the TensorRT baseline degrades substantially, while TMPDiff preserves high fidelity at high speedup. 
On Sana and PixArt, TMPDiff similarly outperforms both baselines by 10 to 20\% across all metrics, though the absolute fidelity is lower due to the more aggressive W4A4 quantization that these models are evaluated under.

\subsection{Mix of Models}
\label{subsec:mixofmodels}

To demonstrate that our TMPDiff framework principle generalizes beyond quantization, we apply it to mixing models of different sizes across denoising steps. 
Specifically, we combine the 1.6B and 600M parameter versions of Sana in a 16+4 step pipeline, using our sensitivity scores to assign the smaller model to the four least sensitive timesteps. 

\Cref{tab:model_mix} shows that TMPDiff outperforms the heuristic of Yang \etal~\cite{yangdenoising}, who place the smaller model at the end of the pipeline based on hand-crafted rules, on all three evaluated metrics. 
Notably, our sensitivity-based schedule only requires linear-time for calibration, while the evolutionary search based approach of~\cite{yangdenoising}, termed Denoising Diffusion Step-aware Models (DDSM), requires costly exploration of the scheduling space and is not scalable to larger models. 
This shows that per-timestep sensitivity scoring captures something fundamental about the denoising process that extends beyond quantization error.

\begin{table}[t]
\centering
\caption{Result of mixing models}
\label{tab:model_mix}
\begin{tabular}{@{}lccc@{}}
\toprule
\textbf{Method} & \textbf{SSIM}$\uparrow$ & \textbf{PSNR}$\uparrow$ & \textbf{LPIPS}$\downarrow$ \\ \midrule
Small Sana     & 0.47 & 13.0 & 0.48  \\ \midrule
DDSM heuristic & 0.80 & 22.7 & 0.09  \\
Ours           & 0.82 & 23.3 & 0.08  \\ \bottomrule
\end{tabular}
\end{table}

\section{Related Work}
\label{sec:relwork}

Extensive research has addressed the quantization of diffusion models \cite{li2023q,li2023qdiff,tang2024post,yao2024timestep,heefficientdm,so2023temporal, ryudgq}, predominantly through \emph{post-training} quantization~\cite{he2023ptqd, ye2025pqd, wang2024towards, shang2023ptqdm}, where pre-trained full-precision models are adapted to integer arithmetic. 
SVDQuant~\cite{li2025svdquant}, the current state-of-the art low bitwidth quantization and our primary quantization method used, achieves practical 4-bit quantization by using low-rank Singular Value Decomposition (SVD) adapters in parallel with quantized layers. 
However, it applies a fixed quantization over all model parameters and timesteps~\cite{zhao2025viditq}, limiting control over quality-efficiency tradeoffs. 
Although \textit{spacial quantization}~\cite{zhao2024mixdq, feng2025mpqdm, kim2025mixdit} addresses this limitation by assigning
different bitwidths to layers, blocks, or channels, these approaches use fixed bitwidths at runtime, disregarding per-timestep error sensitivities under quantization. 

Prior work on error accumulation for quantized diffusion forms our theoretical foundation. 
He \etal~\cite{he2023ptqd} observe the correlation between a quantized layer's input and output, decomposing it into a linear correlation with a normally distributed noise. 
Li \etal~\cite{li2023qdiff} report that the accumulated quantization error converges towards the end of the denoising process without modeling the phenomenon. 
Liu \etal~\cite{liu2025error} derive the closed-form error accumulation we build on in~\cref{sec:explorationoftmpeffects} 
, but without experimental validation across different schedules or models. 

The idea of changing model configuration across the denoising process has been explored in different contexts beyond quantization. 
\citet{liu2023oms,yangdenoising} assign different model sizes per timestep using hand-crafted heuristics or evolutionary search, the latter of which we directly outperform in~\cref{subsec:mixofmodels} with a simpler, linear-time scheduling procedure. 
CacheQuant~\cite{liu2025cachequant}, DeepCache~\cite{ma2023deepcache} and FORA~\cite{selvaraju2024forafastforwardcachingdiffusion} exploit temporal redundancies through activation caching, complementary and combinable as illustrated in \Cref{sec:combined}.

\section{Conclusion}
We explore mixed-precision quantization along the temporal axis of diffusion steps for text-to-image generation. 
We propose an additive error model for temporal mixed-precision diffusion and empirically verify it on four distinct models and three datasets.
The additive model enables temporal precision scheduling as a practical and principled optimization axis for efficient diffusion model inference, complementary to existing spatial mixed-precision approaches. 
We then introduce TMPDiff, a temporal mixed-precision framework with a greedy adaptive bisectioning algorithm for assigning different numeric precisions to different denoising steps. 
Across four diffusion models and three datasets, TMPDiff achieves 10 to 20\% improvement in perceptual quality over uniform-precision baselines at matched speedups, reaching 90\% SSIM on FLUX.1-dev at $2.5\times$ speedup. 

\emph{Limitations:} Efficient kernel-level integration of precision switching is left for future work. 
Weight offloading~\cite{aminabadi2022deepspeed} could reduce the memory overhead of storing both full-precision and quantized weight sets, while nested quantization, i.e. one model supporting both precisions, such as AnyPrecisionLLM~\cite{park2024anyprecision}, suggest a path towards eliminating it entirely. 

\section*{Acknowledgement}
This research has been funded by the Federal Ministry of Education and research of Germany and the state of North Rhine-Westphalia as part of the Lamarr Institute for Machine Learning and Artificial Intelligence.

\bibliographystyle{splncs04nat}
\bibliography{main}

\clearpage
\appendix
\setcounter{page}{1}

%\onecolumn
{
\centering
\Large
% \textbf{\thetitle}\\
\vspace{0.5em}\textbf{Supplementary Material} \\
\vspace{1.0em}
% \twocolumns
% \maketitlesupplementary
}

\section{Validations on Out-of-Distribution samples}
\label{app:outofdistribution}

Having studied the additivity of individual timestep errors in Sec.~\ref{subsec:maineffect} using the same 128 samples for both single-toggle measurements and multi-toggle validation, we verify that these findings generalize to a held-out dataset not used during calibration to reflect the actual deployment scenario.
We evaluate the predictor on 20 random schedules for $K \in \{2, 6, 10, 14, 18\}$ using 128 new samples from the Microsoft Common Objects in COntext (COCO, \cite{lin2014microsoft}) and Densely Captioned Images (DCI, \cite{urbanek2024picture}) datasets each. 

\begin{figure}[h]
\begin{minipage}[h]{0.49\linewidth}
    \centering
    \includegraphics[width=\linewidth]{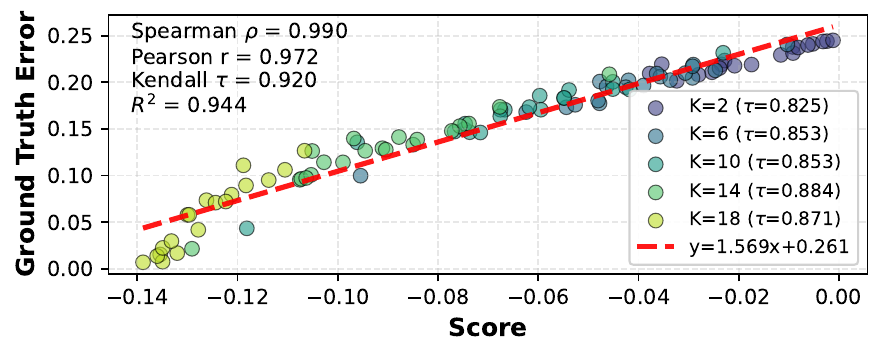}
    \caption{Linear fit between ranking scores $S^{\uparrow}(Z)$ and measured ground truth variance for the Sana model on new COCO samples.}
    \label{fig:prediction_vs_gt_coco_sana}
\end{minipage}
\hfill
\begin{minipage}[h]{0.49\linewidth}
    \centering
    \includegraphics[width=\linewidth]{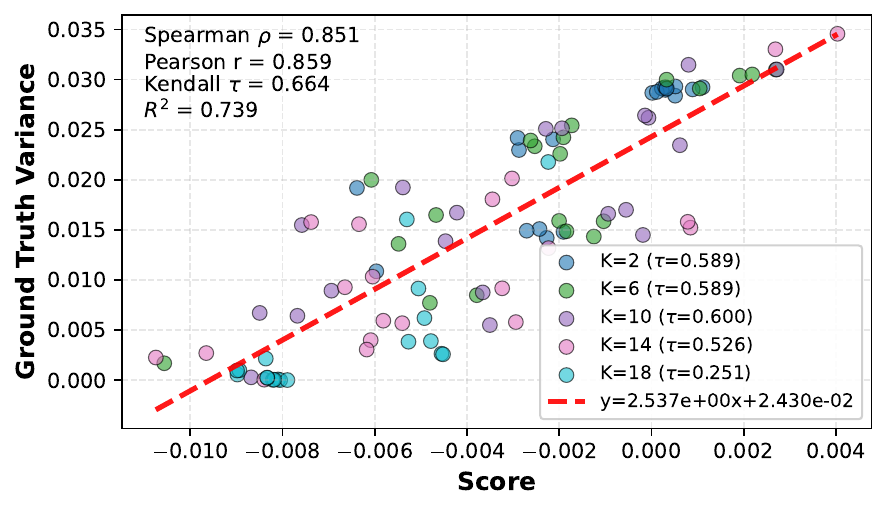}
    \caption{Linear fit between the ranking scores of $S^{\uparrow}(Z)$ and measured ground truth variance for the Sana model on DCI samples.}
    \label{fig:variance_prediction_dci_sana}
\end{minipage}
\end{figure}

\Cref{fig:prediction_vs_gt_coco_sana} shows that the strong correlations observed persists on out-of-distribution samples.
The predictor scores also correlate with directly measured variance as shown in \cref{fig:variance_prediction_dci_sana}, indicating that a schedule optimized for low mean error also produces more consistent results across diverse prompts. 
Hence, schedules performing well on average are also less likely to produce outlier failures. 

\section{Ablations on TMPDiff Calibration}
\label{app:calibrationablation}

\begin{figure*}[h]
    \centering
    \begin{subfigure}[b]{0.49\textwidth}
        \centering
        \includegraphics[width=\linewidth]{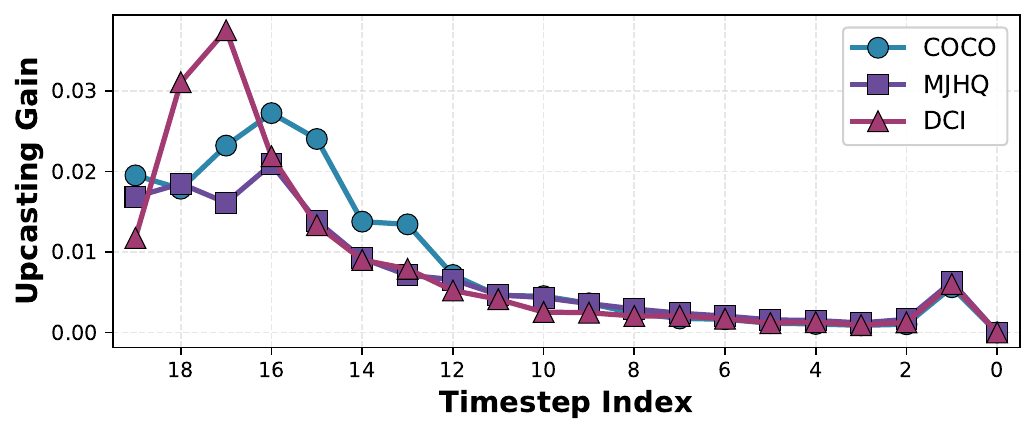}
        \caption{Upcasting gain for samples from different datasets.}
        \label{subfig:app_upcasting_gain_ablation_datasets}
    \end{subfigure}
    \hfill
    \begin{subfigure}[b]{0.49\textwidth}
        \centering
        \includegraphics[width=\linewidth]{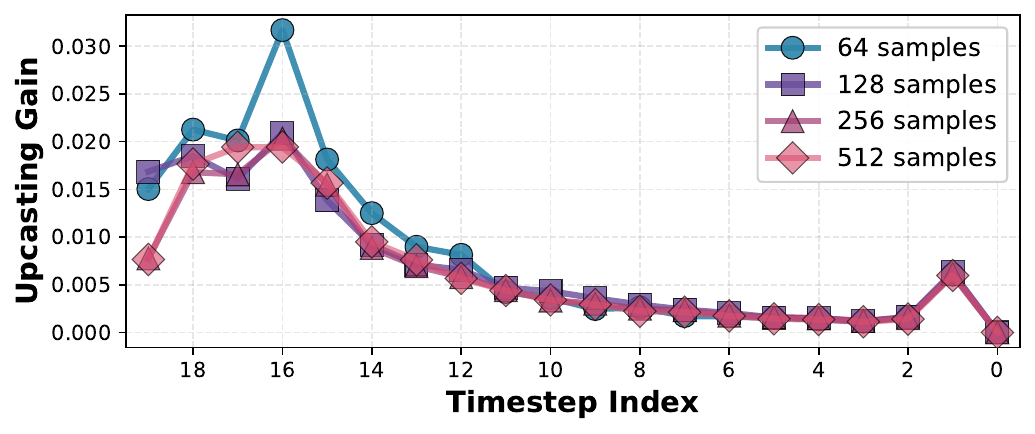}
        \caption{Upcasting gain for different numbers of samples.}
        \label{subfig:app_upcasting_gain_ablation_samples}
    \end{subfigure}
    \caption{Final latent error $\delta_0$ reduction for single-timestep upcasting, with the upcasting gain $\Delta^{\uparrow}_t$ computed on 128 samples of the MJHQ, DCI and COCO datasets respectively (\cref{subfig:app_upcasting_gain_ablation_datasets}) and for 64, 128, 256 and 512 samples from the MJHQ dataset (\cref{subfig:app_upcasting_gain_ablation_samples}) on the Sana model.}
    \label{fig:app_upcasting_gain_ablation}
\end{figure*}

\begin{figure*}[h]
    \centering
    %%%%%%%%% Test = MJHQ %%%%%%%%%
    \begin{subfigure}[b]{0.48\textwidth}
        \centering
        \includegraphics[width=\linewidth]{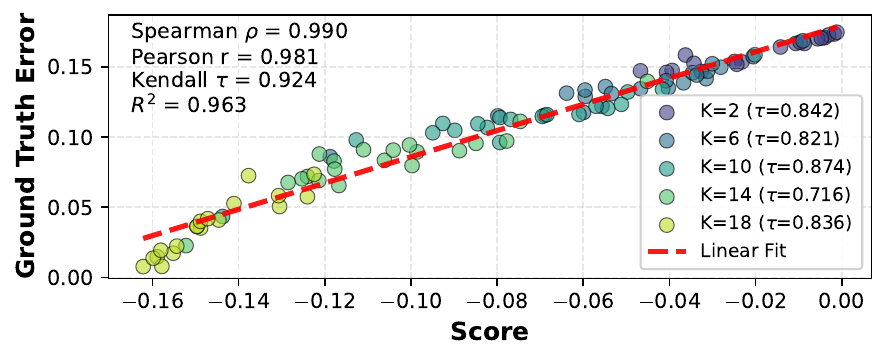}
        \caption{MJHQ test - gains from MJHQ (64 calibration samples)}
    \end{subfigure}\hfill
    \begin{subfigure}[b]{0.48\textwidth}
        \centering
        \includegraphics[width=\linewidth]{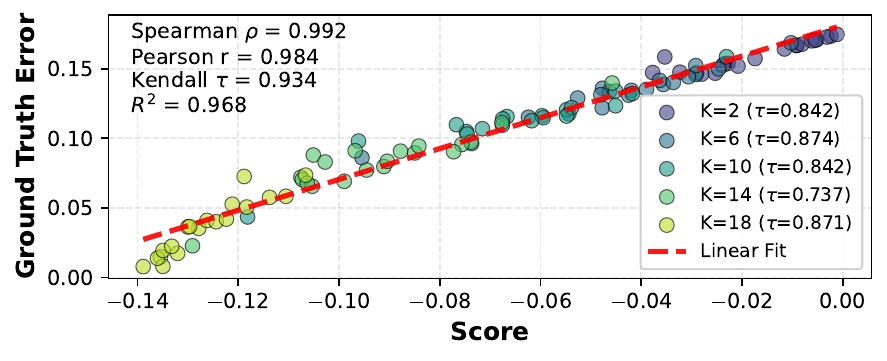}
        \caption{MJHQ test - gains from MJHQ (128 calibration samples)}
    \end{subfigure}

    \vspace{0.7em}
    \begin{subfigure}[b]{0.48\textwidth}
        \centering
        \includegraphics[width=\linewidth]{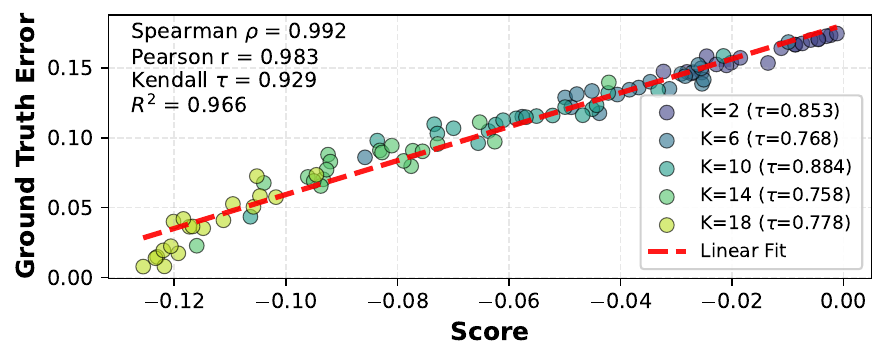}
        \caption{MJHQ test - gains from MJHQ (256 calibration samples)}
    \end{subfigure}\hfill
    \begin{subfigure}[b]{0.48\textwidth}
        \centering
        \includegraphics[width=\linewidth]{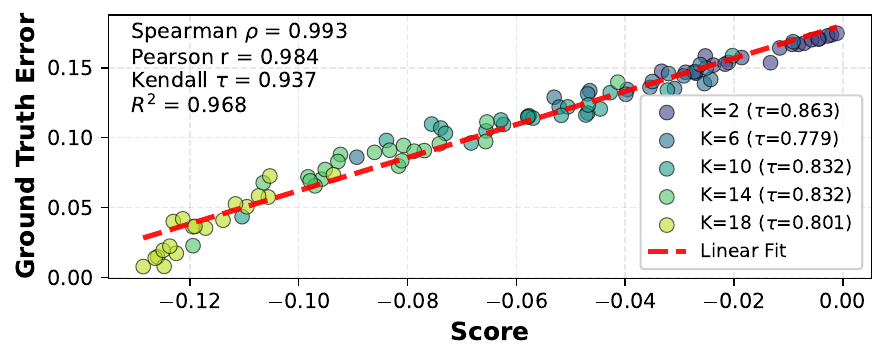}
        \caption{MJHQ test - gains from MJHQ (512 calibration samples)}
    \end{subfigure}

    \caption{Linear fits between \textbf{MJHQ} ground-truth final latent errors $\delta_0$ (128 test samples; 20 schedules per $K\!\in\!\{2,6,10,14,18\}$) and the scoring function  $S^{\uparrow}(Z)$. Upcasting gains $\Delta^{\uparrow}_t$ are computed on \textbf{MJHQ} with $m\in\{64,128,256,512\}$ calibration samples.}
    \label{fig:app_final_error_prediction_mjhq_samplesweep}
\end{figure*}

\begin{figure*}[h]
    \centering
    %%%%%%%%% Test = DCI %%%%%%%%%
    \begin{subfigure}[b]{0.48\textwidth}
        \centering
        \includegraphics[width=\linewidth]{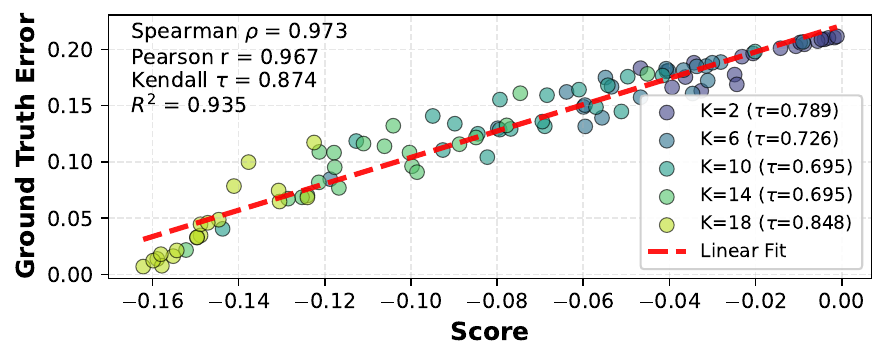}
        \caption{DCI test - gains from MJHQ (64 calibration samples)}
    \end{subfigure}\hfill
    \begin{subfigure}[b]{0.48\textwidth}
        \centering
        \includegraphics[width=\linewidth]{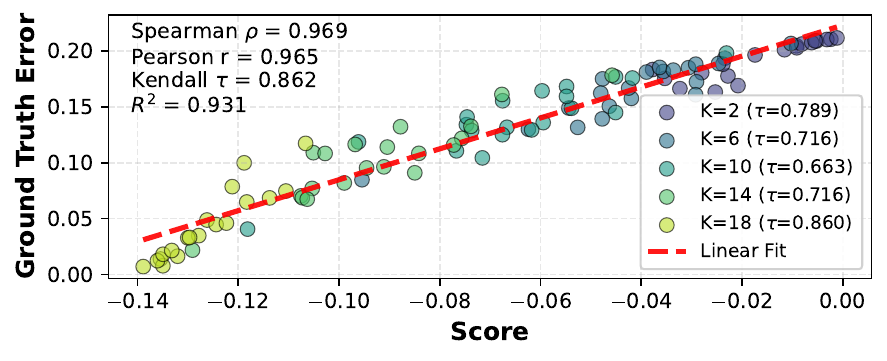}
        \caption{DCI test - gains from MJHQ (128 calibration samples)}
    \end{subfigure}

    \vspace{0.7em}
    \begin{subfigure}[b]{0.48\textwidth}
        \centering
        \includegraphics[width=\linewidth]{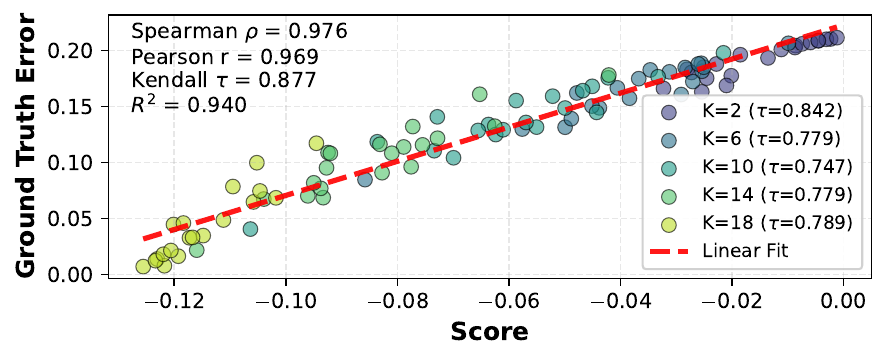}
        \caption{DCI test - gains from MJHQ (256 calibration samples)}
    \end{subfigure}\hfill
    \begin{subfigure}[b]{0.48\textwidth}
        \centering
        \includegraphics[width=\linewidth]{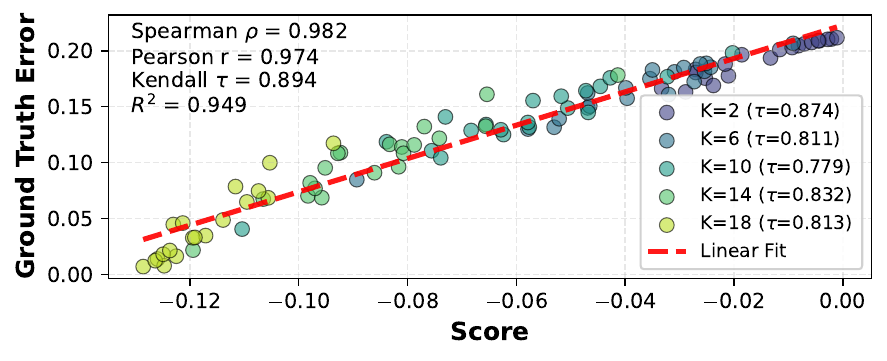}
        \caption{DCI test - gains from MJHQ (512 calibration samples)}
    \end{subfigure}

    \caption{Linear fits between \textbf{DCI} ground-truth final latent errors $\delta_0$ (128 test samples; 20 schedules per $K\!\in\!\{2,6,10,14,18\}$) and the scoring function  $S^{\uparrow}(Z)$. Upcasting gains $\Delta^{\uparrow}_t$ are computed on \textbf{MJHQ} with $m\in\{64,128,256,512\}$ calibration samples.}
    \label{fig:app_final_error_prediction_dci_samplesweep}
\end{figure*}

\begin{figure*}[h]
    \centering
    %%%%%%%%% Test = COCO %%%%%%%%%
    \begin{subfigure}[b]{0.48\textwidth}
        \centering
        \includegraphics[width=\linewidth]{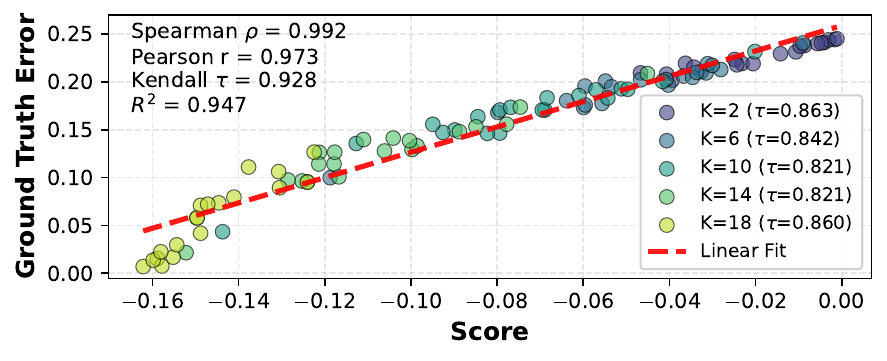}
        \caption{COCO test - gains from MJHQ (64 calibration samples)}
    \end{subfigure}\hfill
    \begin{subfigure}[b]{0.48\textwidth}
        \centering
        \includegraphics[width=\linewidth]{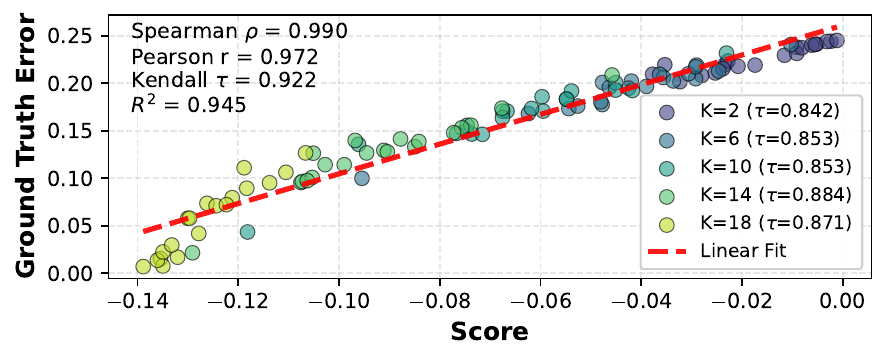}
        \caption{COCO test - gains from MJHQ (128 calibration samples)}
    \end{subfigure}

    \vspace{0.7em}
    \begin{subfigure}[b]{0.48\textwidth}
        \centering
        \includegraphics[width=\linewidth]{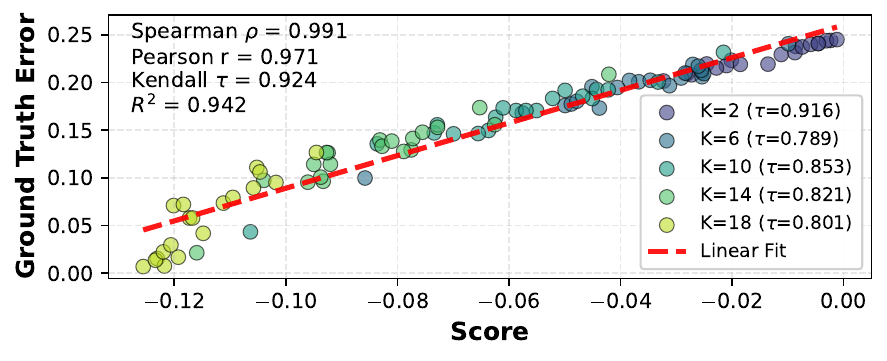}
        \caption{COCO test - gains from MJHQ (256 calibration samples)}
    \end{subfigure}\hfill
    \begin{subfigure}[b]{0.48\textwidth}
        \centering
        \includegraphics[width=\linewidth]{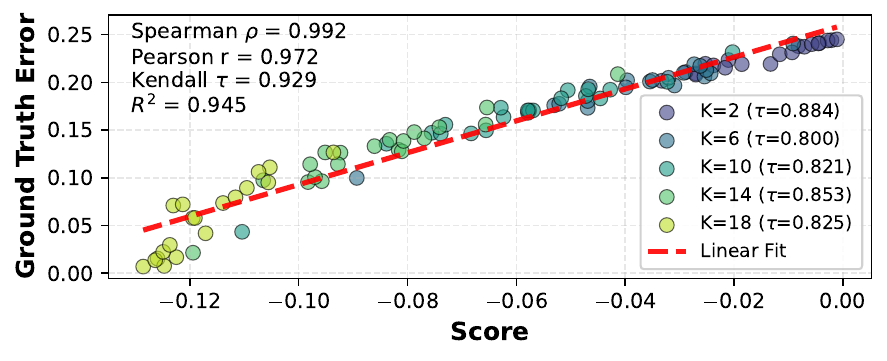}
        \caption{COCO test - gains from MJHQ (512 calibration samples)}
    \end{subfigure}

    \caption{Linear fits between \textbf{COCO} ground-truth final latent errors $\delta_0$ (128 test samples; 20 schedules per $K\!\in\!\{2,6,10,14,18\}$) and the scoring function  $S^{\uparrow}(Z)$. Upcasting gains $\Delta^{\uparrow}_t$ are computed on \textbf{MJHQ} with $m\in\{64,128,256,512\}$ calibration samples.}
    \label{fig:app_final_error_prediction_coco_samplesweep}
\end{figure*}

We conduct ablations on how the number of calibration samples and their source used to compute the upcasting gain $\Delta_t^{\uparrow}$ impacts ranking. 
\Cref{fig:app_upcasting_gain_ablation} illustrates that samples from different datasets and different amounts of calibration samples impact which timesteps rank highest. 
Since the ground-truth order of timesteps depends on the data distribution, obtaining it for large and diverse datasets is infeasible. 
Hence, with a finite calibration set, we try to capture tendencies, not a unique universal order. 
Despite such differences, the scoring functions $S^{\uparrow}(Z)$ constructed using different amounts of calibration samples $m\in\{64,128,256,512\}$ all show a strong linear fit when tested on different datasets as shown in \Cref{fig:app_final_error_prediction_mjhq_samplesweep,fig:app_final_error_prediction_dci_samplesweep,fig:app_final_error_prediction_coco_samplesweep}. 
For our main deployment setting with $K=5$ full-precision timesteps, the practical impact of calibration differences diminishes. 
Once the top-5 most sensitive timesteps are captured, their internal ordering does not matter for schedule quality anymore. 
Therefore, we adopt $m=128$ calibration samples in our experiments for more stability over 64 samples, but refrain from using more samples due to diminishing returns. 

\section{Latent Errors as Image Metric Proxy}
\label{app:latenterrorproxy}

Since image-space metrics (FID, LPIPS, CLIP Score) require large sample sizes to produce stable estimates~\cite{chong2019effectively}, making them unreliable for comparing individual schedules during calibration, we use the final latent error $E(Z)$ as a proxy throughout this paper. 
This choice is supported by its strong correlation with perceptual metrics as shown in~\cref{tab:correlation_metrics}.

\begin{table}[h]
    \centering
    \caption{Pearson ($r_p$) and Spearman ($r_s$) correlation coefficients between image quality metrics and final latent errors.}
    \label{tab:correlation_metrics}
    \small  % Makes font slightly smaller
    \begin{tabular}{@{}llccccc@{}}
        \toprule
        \textbf{Model} & & \textbf{IR}$\uparrow$ & \textbf{CLIPS}$\uparrow$ & \textbf{LPIPS}$\downarrow$ & \textbf{SSIM}$\uparrow$ & \textbf{PSNR}$\uparrow$ \\
        \midrule
        \multirow{2}{*}{PixArt} 
            & $r_p$   & -0.69 & -0.74 & 0.73 & -0.59 & -0.92 \\
            & $r_s$   & -0.72 & -0.78 & 0.82 & -0.67 & -0.93 \\
        \addlinespace[0.5em]
        \multirow{2}{*}{Sana} 
            & $r_p$   & -0.67 & -0.42 & 0.84 & -0.67 & -0.74 \\
            & $r_s$   & -0.72 & -0.41 & 0.90 & -0.74 & -0.87 \\
        \bottomrule
    \end{tabular}
\end{table}

\section{Mixed Memory} \label{app:memtrans}

Once the optimal schedule  $Z^{\ast}$ is obtained, we use it to switch between the full-precision and the quantized model between denoising steps during inference. 
The easiest way is to have both models stored in memory at the same time, causing no latency overhead at all. 
Otherwise, reported speedups must assume that the precision-switching overhead can be hidden at the hardware level. 
However, efficient kernel-level integration remains an open engineering problem that is left for future work. 
Weight offloading~\cite{aminabadi2022deepspeed}, which offloads the weights of the current model as soon as they are no longer needed and the weights of the incoming model are streamed in ahead of the precision switch, represents a natural direction for realizing TMPDiff without requiring both weight sets to be stored in GPU memory simultaneously. 
Alternatively, nested quantization~\cite{park2024anyprecision}, which derives lower-precision representations directly from shared higher-precision parameters, could eliminate the need for storing multiple models entirely. 

\begin{figure}[h]
    \centering
    \includegraphics[width=0.5\linewidth]{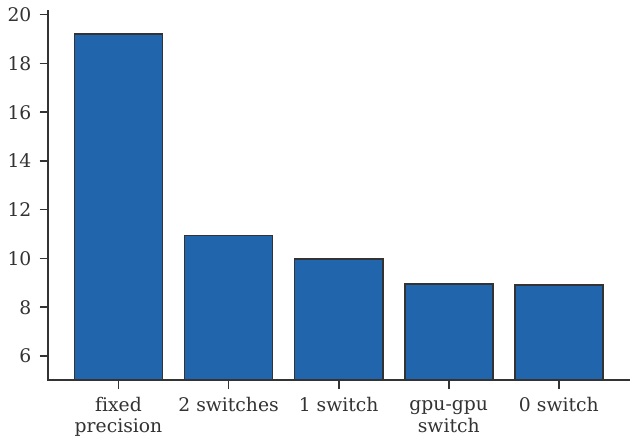}
    \caption{Sampling speed (seconds) for FLUX on several memory transfer strategies.}
    \label{fig:memtransfer}
\end{figure}

We report in \cref{fig:memtransfer} the impact on latency of keeping only one model in memory. Sampling with 15 quantized and 5 floating point steps, we consider the case where both models remain on the GPU (0 switch) and the cases where we have to swap the model once (1 switch) or twice (2 switches). We additionally consider the case where we keep one separate GPU for each model and switch the noisy latent during inference instead (gpu-gpu switch). The results shows switching the memory shows a limited overhead, not comparable to the cost of using a full-precision model.

\section{Evaluation on Reference-free Metrics}
\label{app:tables}

In the main body we assess the impact of TMPDiff on output fidelity by using the full-precision model as reference, reporting PSNR, SSIM and LPIPS. 
Reference-free metrics are less discriminative in our controlled setting but are reported in~\cref{tab:gt_results} as corroborative evidence for improved image quality over uniform precision baselines at matched end-to-end speedup. 

\begin{table}[h]
\centering
\caption{Comparison of Image Reward, CLIP Score, and FID (\wrt ground truth images) across datasets. The model evaluated is Sana, using identical settings to the experiment reported in \cref{tab:results_everything}. }
\label{tab:gt_results}
\begin{tabular}{l l|ccc}
\toprule
\textbf{Dataset} & \textbf{Metric} & \textbf{Ours} & \textbf{bfloat16} & \textbf{w4a4r32} \\
\midrule
\multirow{3}{*}{\textbf{COCO}} 
 & IR ($\uparrow$) & \textbf{1.0434} & 0.9476 & 1.0413 \\
 & CLIP ($\uparrow$) & \textbf{27.021} & 26.925 & 26.907 \\
 & FID ($\downarrow$) & \textbf{71.243} & 71.548 & 71.839 \\
\midrule
\multirow{3}{*}{\textbf{MJHQ}} 
 & IR ($\uparrow$) & \textbf{1.0438} & 0.9239 & 1.0424 \\
 & CLIP ($\uparrow$) & 27.163 & 26.971 & \textbf{27.172} \\
 & FID ($\downarrow$) & \textbf{55.990} & 57.196 & 55.947 \\
\midrule
\multirow{3}{*}{\textbf{sDCI}} 
 & IR ($\uparrow$) & 1.0155 & 0.8131 & \textbf{1.0172} \\
 & CLIP ($\uparrow$) & \textbf{26.728} & 26.540 & 26.707 \\
 & FID ($\downarrow$) & \textbf{62.468} & 67.037 & 62.447 \\
\bottomrule
\end{tabular}
\end{table}

\section{Evaluation and System Setup}
\label{app:testbed}

Experiments were run on Ubuntu 22.04 with Python 3.12 and Pytorch 2.7 for CUDA 12.6. The hardware platform consists of one Nvidia RTX 4090 GPU with one Intel Core i7-14700KF CPU. The results reported in \cref{tab:gt_results,tab:results_everything} were run on 1024 samples of each dataset, generated with a 1024$\times$1024 resolution with guidance scale 4.5. We additionally ran PixArt experiments with 5,000 images as a sanity check, but they did not show any significant difference in the evaluation results. 
All speedups are computed over 16 end-to-end backbone inference runs following 10 warmup iterations, excluding the text encoding and image decoding stages.

\section{Orthogonality of TMPDiff to Caching-Based Acceleration} \label{sec:combined}
An additional benefit of using temporal mixed quantization as an acceleration vector is that it maintains the ability to set up other classic acceleration frameworks. To illustrate, we consider the results of running TMPDiff on a DeepCache-accelerated sampler. DeepCache~\cite{ma2023deepcache} is an acceleration technique that skips parts of the denoising process by reusing features from previous steps.

We report in \cref{tab:combined} the experimental results of using the different methods on SDXL with the MJHQ dataset. We observe that both methods provide a significant increase in sampling speed with a slight decrease in image quality. With our chosen setup, the trade-off leans more towards speed for DeepCache. The results when the two approaches are combined show a substantial increase in speed (+22\% \wrt DeepCache only) with minimal generation decay. This confirms TMPDiff can be used seamlessly with existing acceleration frameworks without requiring adaptation or specific setups.
 
\begin{table}[h]
\centering
\caption{Result of combined DeepCache and TMPDiff sampling}
\label{tab:combined}
\begin{tabular}{@{}cccccc@{}}
\toprule
\textbf{DeepCache} & \textbf{TMPDiff}  &  \textbf{SSIM}$\uparrow$ & \textbf{PSNR}$\uparrow$ & \textbf{LPIPS}$\downarrow$  & \textbf{Speedup}\\ \midrule
\checkmark & \ding{55}  & 0.73 & 20.5 & 0.28 & 2.25  \\
\ding{55}  & \checkmark & 0.85  & 28.0  & 0.10 & 1.42   \\
\checkmark & \checkmark & 0.72 & 20.3 & 0.29  & 2.91 \\ \bottomrule
\end{tabular}
\end{table}

% TODO FINAL: WARNING: do not forget to delete the supplementary pages from your submission 

\end{document}